\documentclass{article}

\usepackage{arxiv}

\usepackage[utf8]{inputenc} 
\usepackage[T1]{fontenc}    
\usepackage{hyperref}       
\usepackage{url}            
\usepackage{booktabs}       
\usepackage{amsfonts}       
\usepackage{nicefrac}       
\usepackage{microtype}      

\usepackage{lipsum}         
\usepackage{graphicx}
\usepackage{natbib}
\usepackage{doi}
\usepackage{amsmath}
\usepackage{amssymb}
\usepackage{mathtools}
\usepackage{amsthm}
\usepackage{multirow}

\usepackage{cleveref}       

\def \equref #1{Eq.~(\ref{#1})}

\def \secref #1{Section~\ref{#1}}
\def \figref #1{{Figure~\ref{#1}}}

\title{Self-Supervised Learning for Group Equivariant Neural Networks}

\date{}

\author{Yusuke Mukuta\\
	The University of Tokyo / RIKEN\\
	\texttt{mukuta@mi.t.u-tokyo.ac.jp} \\
	\And
        Tatsuya Harada\\
	The University of Tokyo / RIKEN\\
	\texttt{harada@mi.t.u-tokyo.ac.jp} \\
}

\renewcommand{\shorttitle}{Self-Supervised Learning for Group Equivariant Neural Networks}

\hypersetup{
pdftitle={Self-Supervised Learning for Group Equivariant Neural Networks},
pdfsubject={cs.AI, cs.CV},
pdfauthor={Yusuke Mukuta, Tatsuya Harada},
pdfkeywords={Self-supervised learning, Group equivariance},
}

\begin{document}
\maketitle

\begin{abstract}
  This paper proposes a method to construct pretext tasks for self-supervised learning on group equivariant neural networks. Group equivariant neural networks are the models whose structure is restricted to commute with the transformations on the input. Therefore, it is important to construct pretext tasks for self-supervised learning that do not contradict this equivariance. To ensure that training is consistent with the equivariance, we propose two concepts for self-supervised tasks: equivariant pretext labels and invariant contrastive loss. Equivariant pretext labels use a set of labels on which we can define the transformations that correspond to the input change. Invariant contrastive loss uses a modified contrastive loss that absorbs the effect of transformations on each input. Experiments on standard image recognition benchmarks demonstrate that the equivariant neural networks exploit the proposed equivariant self-supervised tasks.
\end{abstract}


\section{Introduction}
Self-supervised learning is the method by which we define pretext tasks based on our prior knowledge about the input data and train the feature extractor using the pretext tasks without supervised labels.
Pretext tasks include tasks to solve ill-posed problems such as image completion, tasks to predict image context, and tasks to make the features of the augmented images from the same image close to each other.
Currently, self-supervised learning demonstrates comparative accuracy to supervised learning and is an effective framework for learning features in an unsupervised manner.

Another direction for utilizing prior knowledge is to incorporate the knowledge into the model structure of the feature extractor.
For example, the convolutional layer is designed to be robust to the local translation of the object in an image.
Group equivariant neural networks are the effective framework for utilizing the knowledge of invariance for the required transformations such as image rotations and image flipping in the neural networks structure.
Given the input data $x$, the transformation on the input $T_{\mathrm{in}}(g)$, and the transformations on the output $T_{\mathrm{out}}(g)$, the group equivariant neural networks $f_{\theta}$ are constructed to satisfy
$T_{\mathrm{out}}(g)(f_{\theta}(x))=f_{\theta}(T_{\mathrm{in}}(g)(x))$ for any transformation parameter $g$, input data $x$, and model parameter $\theta$.
Because the structure of group equivariant neural networks are restricted to satisfy equivariance, this restriction regularizes the model, and the learned model generally demonstrates better accuracy than the standard non-equivariant neural networks.

Therefore, we expect that we can obtain the effective feature learning method by combining these two ideas to utilize the prior knowledge.

When we combine the idea of self-supervised learning and the group equivariant neural networks, it is important to design the method such that these two components do not adversely affect each other.
The functions that the group equivariant neural networks can learn are restricted to the mappings that preserve equivariance. Therefore, when a pretext task requires the mapping to violate the equivariance, it is difficult to learn the function through this pretext task.

In this paper, we propose a self-supervised task that is suitable for group equivariant neural networks.
The idea is to construct a self-supervised loss that does not change under the transformations on the input data.
This invariance guarantees that we can learn the same equivariant model even when we apply the considered transformations to the input.
To construct a self-supervised loss that satisfies this condition, we propose two concepts for self-supervised loss: equivariant pretext labels and invariant contrastive loss.
Equivariant pretext labels are constructed such that they are consistent with the considered transformations on the input. This indicates that when we apply the transformations on the input, the corresponding pretext labels are also changed according to the considered transformations. The invariant contrastive loss is a loss that does not change when we apply transformations to each input data.
We extend several existing self-supervised pretext tasks to satisfy these concepts.

We apply the proposed loss to the image classification model on the ImageNet dataset and evaluate the model using several image recognition benchmarks.
We demonstrate that the trained group equivariant neural networks demonstrate good classification accuracy when we use the proposed loss.

The contributions of the paper are as follows:
\begin{itemize}
\item We propose the concepts of equivariant pretext labels and invariant contrastive loss to train the group equivariant neural networks in a self-supervised manner.
\item We propose an equivariant extension to several existing self-supervised tasks.
\item We apply our method to standard image recognition benchmarks and demonstrate the effectiveness of the proposed loss.
\end{itemize}

\section{Related Work}
\subsection{Self-Supervised Learning}
Self-supervised learning is one of the frameworks used to train the feature extractor in an unsupervised manner.
In self-supervised learning, we first design a task to minimize the loss function based on the prior knowledge of the input data and pretrain the model with the task.
Next, we fine-tune the model initialized with the pretrained parameter on the target dataset with labels to obtain the classifier.
In image recognition, early research on self-supervised learning is based on the hand-crafted pretext task. We design an ill-conditioned problem and expect the model to acquire basic knowledge about the image through training to solve the problem.
Some examples of pretext tasks are tasks to recover an input image from the image with incomplete information \citep{pathak2016context,zhang2016colorful,zhang2017split,he2022masked}, tasks to predict spatial relationships between subregions of an image, \citep{doersch2015unsupervised,noroozi2016unsupervised,noroozi2018boosting,kim2018learning}, tasks to predict transformations from the transformed input images \citep{komodakis2018unsupervised,zhang2019aet} and tasks to predict image primitives in an image \citep{noroozi2017representation,gidaris2020learning}.

A recent trend in self-supervised learning is to utilize relationships between the images in the dataset instead of considering one-to-one relationships between the images and pretext labels.
In \cite{caron2018deep}, the model is trained iteratively by first assigning a pseudo-label to each image by applying clustering to the output of the current model, and then updating the model to predict the assigned pseudo-label more accurately.
In \cite{asano2019self}, this method is extended by imposing a uniformness restriction to the pseudo-label space and formulating the learning loss as the Kullback-Leibler divergence between the pseudo-label distribution and predicted label distribution.
In \cite{hjelm2018learning,oord2018representation}, the model is trained to maximize mutual information between the intermediate features. These methods calculate loss by utilizing the other images in the dataset as negative samples.
Among data-driven methods, the self-supervised framework based on contrastive loss is well investigated. In contrastive learning, a positive image pair is constructed by applying data augmentation to the same input image and the negative image pair is constructed by sampling different images from the dataset. The model is then trained such that the positive pair is close while the negative pair is distant based on distance criterions.
\cite{chen2020simple} uses the softmax cross-entropy loss for this purpose. In \cite{he2020momentum}, the author introduced a momentum encoder that is obtained from the moving average of the learned feature extractors and the memory bank to preserve the extracted features from the previous mini-batches in the queue data structure.
Variants of contrastive learning include methods to use the consistency between the pseudo label assignment by applying clustering \citep{li2020prototypical,caron2020unsupervised}, methods to match the feature only between the positive pair instead of the softmax cross-entropy loss \citep{grill2020bootstrap,chen2021exploring} and methods that trains the model to output the correlation matrix close to the identity matrix \citep{zbontar2021barlow}.

\subsection{Group Equivariant Neural Networks}
Group equivariant neural networks are the framework for utilizing knowledge about invariance into the model structure of the neural networks.
Given the transformations on the input $T_{\mathrm{in}}(g)$ and output $T_{\mathrm{out}}(g)$ for the group element $g\in G$, group equivariant neural networks $f_{\theta}$ with learnable parameter $\theta$ are constructed to satisfy
$T_{\mathrm{out}}(g) f_{\theta} (x)= f_{\theta} (T_{\mathrm{in}}(g) (x))$ for any input $x$, transformation group element $g$, and the model parameter $\theta$.
Because neural networks are constructed as the composition of layers, we generally construct the group equivariant neural networks by first designing layers that satisfy equivariance, and then composing the equivariant layers.
We introduce the group equivariant neural networks for image recognition in this section.
Group Equivariant CNNs (G-CNNs) \citep{cohen2016group} is the standard model for the group equivariant neural networks. G-CNNs has $|G|$ learnable convolution layers and calculates the layer output by applying group convolution to the input. The first layer calculates the output as:
\begin{equation}
  x^{\mathrm{out}}_g = (T_{w_0}(g) w_0) * x^{\mathrm{in}},\label{eq:equivariantconv0}
\end{equation}
and succeeding layers calculate the output as
\begin{equation}
  x^{\mathrm{out}}_g = \sum_{h \in G} w_{(g h^{-1})} * x^{\mathrm{in}}_h,\label{eq:equivariantconv}
\end{equation}
where $x^{\mathrm{in}}_g$ and $x^{\mathrm{out}}_g$ denote the $g$-th elements of the input and output respectively, $w_0, w$ are the learnable weights, $T_{w_0}$ denotes the corresponding transformation on the weights for the first layer, and $*$ indicates the image convolution.
When we apply $T_{\mathrm{in}}(g)$ to the input $x$, the corresponding output becomes the permutation between the $g$-th elements of $x^{\mathrm{out}}_g$. Therefore, the network becomes group equivariant. 
While G-CNNs are a versatile framework, this method has the drawback that the model carries extensive parameter and computation costs.
Several methods for reducing these costs are proposed.
In \cite{cohen2017steerable}, the computation cost is reduced by first applying the Fourier transformations to the input to decompose it into irreducible representations, and then applying the layer on this irreducible feature space.
\cite{weiler2018learning} represents a group convolutional filter using the weighted sum of the basis filter, and uses the summation coefficient as the learnable parameter to reduce the number of model parameters.
As a different construction, \cite{marcos2017rotation} constructs the rotation equivariant layer by applying the convolution layer while rotating and calculates the two-dimensional vector from the statistics of the activation.
\cite{worrall2017harmonic} constructs the equivariant layer for the two-dimensional continuous rotation using spherical harmonics.
\cite{jenner2022steerable} constructs the convolutional weights by incorporating the weighted sum of the differentiate operator.
These methods and their variants are summarized in \cite{e2cnn}.

\subsection{Self-Supervised Learning with Equivariance}
In self-supervised methods, prior knowledge about invariance is used to construct the loss function rather than incorporate it into the model structure.
For example, in contrastive learning, the invariance information is encoded as the data augmentation and used to construct the positive pair.
\cite{misra2020self} tries to learn the invariance using the similarity between the augmented samples.
\cite{mitrovic2020representation} utilizes the concept of the causal mechanism and explicitly adds the regularizer such that the output features from the same image with different data augmentations resemble each other.
\cite{dangovski2021equivariant} proposes a hybrid method of the pretext tasks to predict the image transformations and contrastive loss to obtain the invariant features under the transformations.
Based on the hybrid framework, \cite{dangovski2021equivariant} analyzes the invariance and equivariance information that contribute to the performance.
\cite{keller2022homomorphic} is close to our work in that the problem of training the group equivariant neural networks is also considered. Although \cite{keller2022homomorphic} constructs the invariant loss by averaging the loss with respect to the augmented features under the group action while the shape of the similarity metric itself is preserved, we directly modify the similarity function such that the loss is instance-wise invariant. Further, \cite{keller2022homomorphic} applies the method to SimCLR \citep{chen2020simple}, while we apply our method to several contrastive and non-contrastive self-supervised tasks.

\begin{figure}[t]
    \begin{minipage}[b]{0.49\hsize}
  \centering
  \includegraphics[width=\hsize]{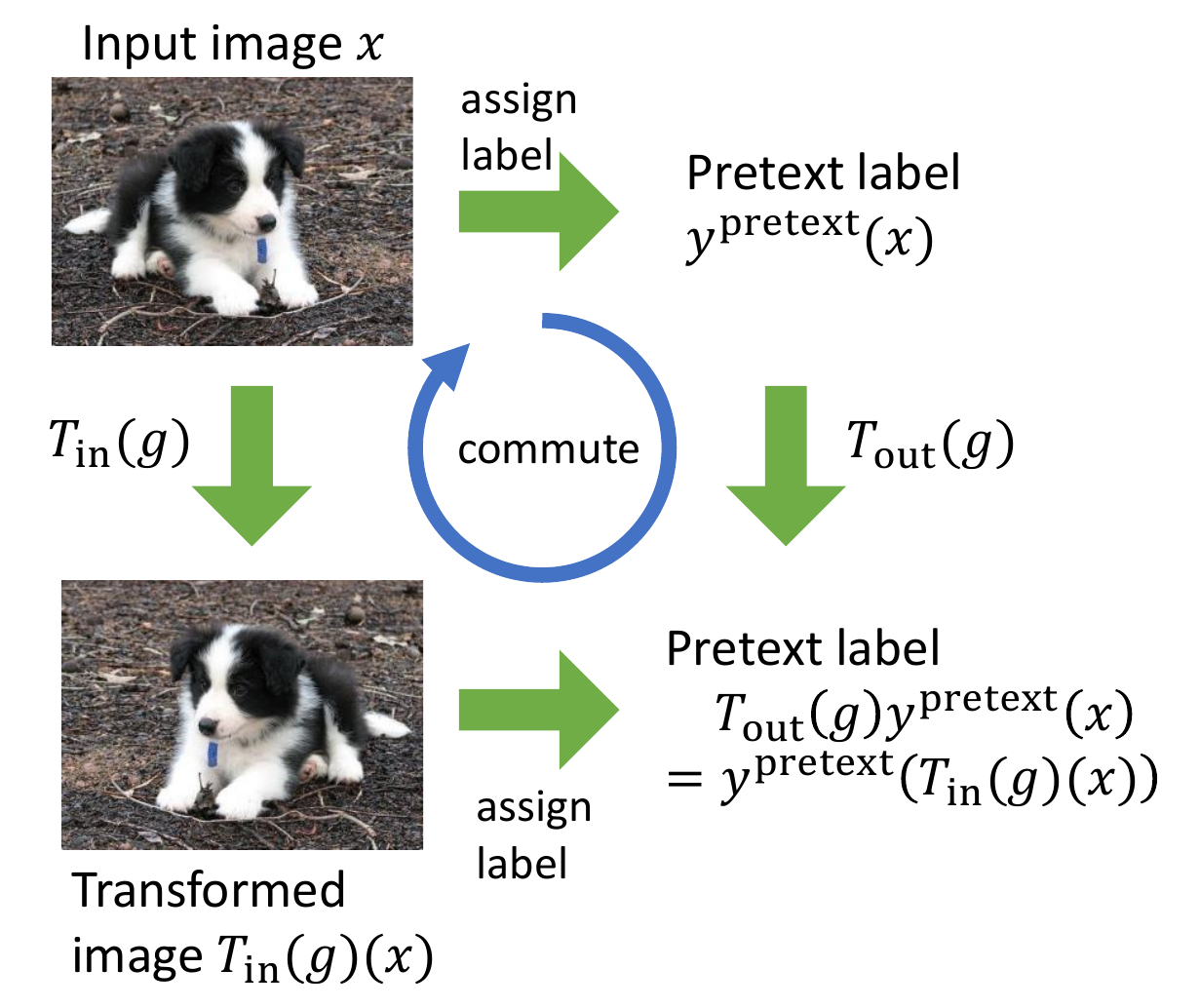}
  \caption{Illustrative image of the equivariant pretext labels.}\label{fig:pretext}
  \end{minipage}
\begin{minipage}[b]{0.49\hsize}
  \centering
  \includegraphics[width=\hsize]{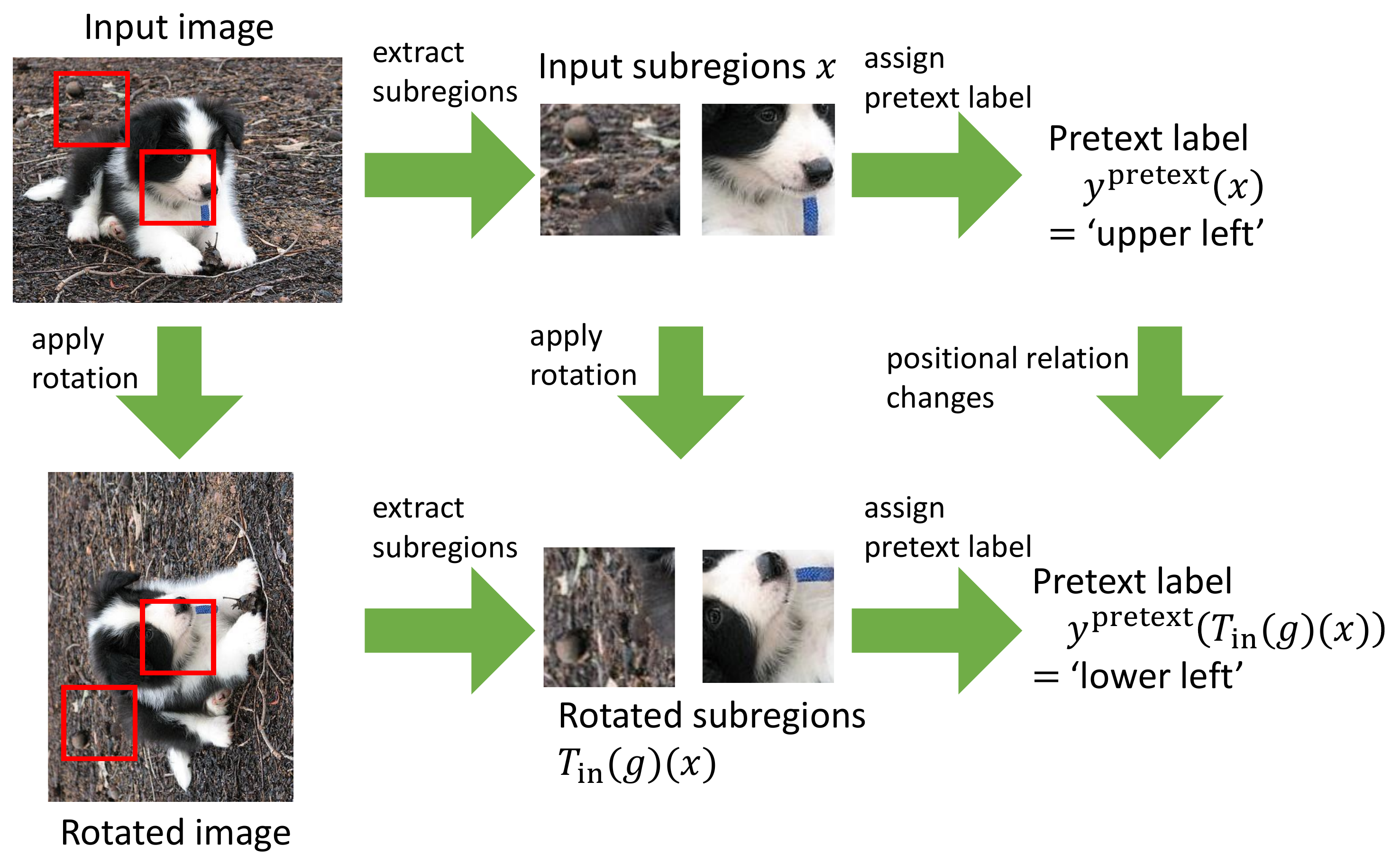}
  \caption{Illustrative image of equivariant context prediction.}\label{fig:context}
    \end{minipage}
\end{figure}

\section{Proposed Method}
\subsection{Motivation}
Our goal is to construct a self-supervised loss that is suitable for the group equivariant neural networks.
We design the loss function to learn $\theta$ for the feature extractor $f_{\theta}$ that satisfies 
$T_{\mathrm{out}}(g) f_{\theta} (x)= f_{\theta} (T_{\mathrm{in}}(g) (x))$.

We first discuss the problem that occurs when we employ the standard loss function. We consider the pretext task where we assign the pretext label $y^{\mathrm{pretext}}(x)$ for the input $x$, and train the model to predict the label $y^{\mathrm{pretext}}(x)$ from $x$. In this case, $f_{\theta}$ is trained to map $x$ onto $y^{\mathrm{pretext}}(x)$, but from the equivariance of $f_{\theta}$, $T_{\mathrm{in}}(g)(x)$ is mapped onto $T_{\mathrm{out}}(g)(y^{\mathrm{pretext}}(x))$.
However, in general, the pretext label for the input $T_{\mathrm{in}}(g)(x)$ denoted as  $y^{\mathrm{pretext}}(T_{\mathrm{in}}(g)(x))$ is not the same as $T_{\mathrm{out}}(g)(y^{\mathrm{pretext}}(x))$.
Therefore, $f_{\theta}$ is trained to map $T_{\mathrm{in}}(g)(x)$ to the multiple target label, and it is difficult to train the model consistently.
This discussion indicates the importance of designing a self-supervised task that is consistent with the group equivariance.

\subsection{Equivariant Pretext Labels}
In the previous section, we discussed the problem that we cannot train the model to mimic one-to-one mapping between input and output when we use standard self-supervised pretext labels.
Therefore, we propose to use the pretext labels that are restricted to reflect equivariance.
We consider the case where the transformation group acts on the label space of $y^{\mathrm{pretext}}$ satisfies
\begin{equation}
y^{\mathrm{pretext}}(T_{\mathrm{in}}(g)(x))=T_{\mathrm{out}}(g)(y^{\mathrm{pretext}}(x)).\label{eq:pretext}
\end{equation}
When this condition is satisfied, the loss function does not change even when we apply $T_{\mathrm{in}}(g)$ to the input $x$, and the training becomes consistent.
In this study, we consider the case where $T_{\mathrm{out}}(g)$ acts as the permutation between labels.
We plot the illustrative image of the equivariant pretext labels in \figref{fig:pretext}.

Below, we propose this equivariant extension of the existing pretext tasks.

\subsubsection{Context prediction with $\pi/2$ rotation}
Context prediction \citep{doersch2015unsupervised} is the method that we first extract $3\times 3$ adjacent image subregions and then use the pair of central subregions and one of the other subregions as the input and train the model to predict the relative position between the input subregions as the eight-category classification problem.
We consider the effect of the $\pi/2$ image rotation on this pretext label.
As plotted in \figref{fig:context}, the rotation acts as the permutation on the relative position label.
For example, the subregion which is upper-left in the original image moves to the lower-left after the rotation, and the right subregion moves to the up subregion after the rotation
Therefore, when we align the relative position category as ['left', 'down', 'right', 'up', 'upper left', 'lower left', 'lower right', 'upper right'], this space becomes the direct sum of two four-dimensional space where $\pi/2$ acts as the permutation.
We define this method to train the equivariant model with this label order as the equivariant context prediction.

\begin{figure}[t]
  \begin{minipage}[b]{0.49\hsize}
  \centering
  \includegraphics[width=\hsize]{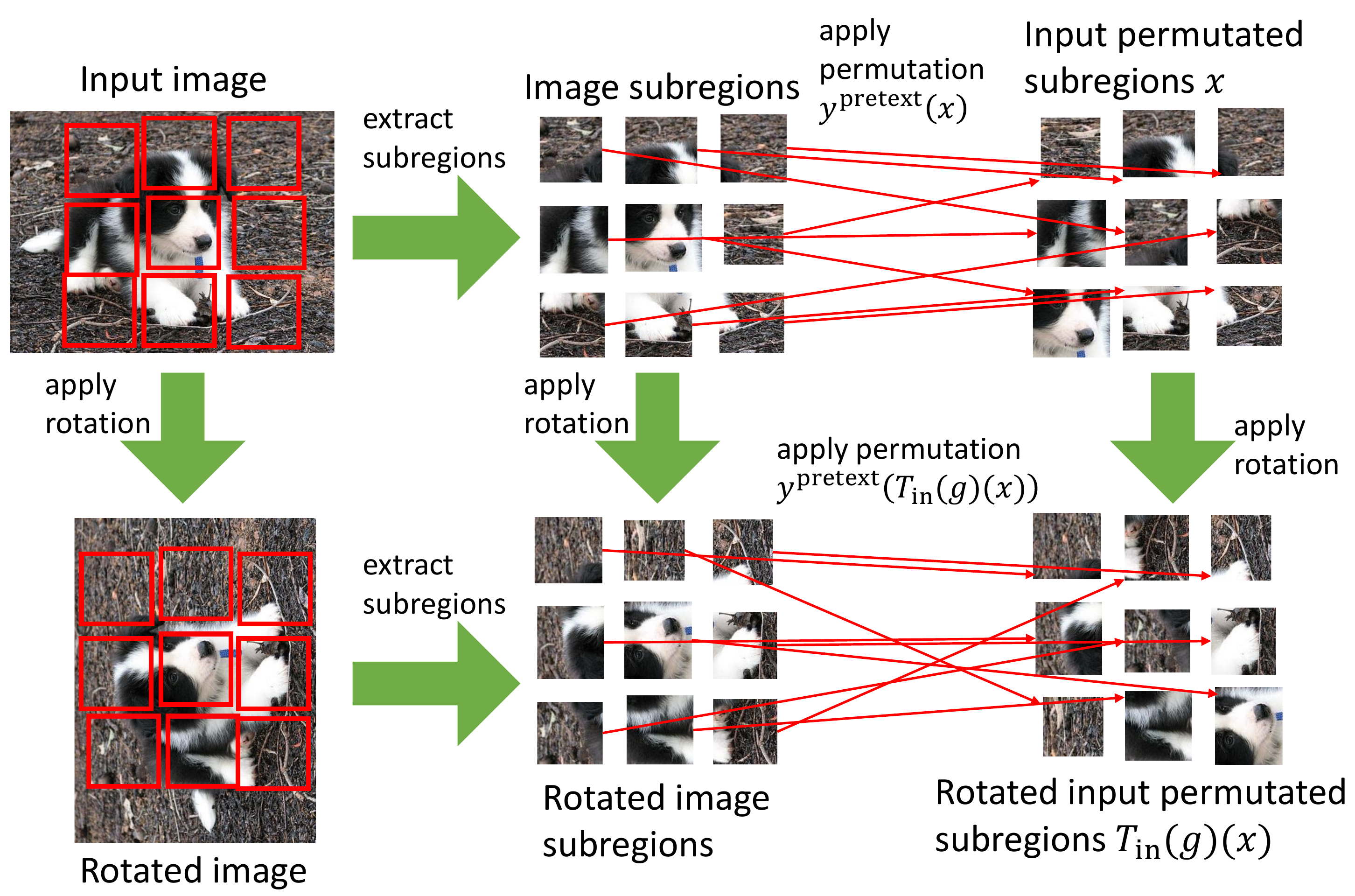}
  \caption{Illustrative image of equivariant jigsaw.}\label{fig:jigsaw}
\end{minipage}
\begin{minipage}[b]{0.49\hsize}
  \centering
  \includegraphics[width=\hsize]{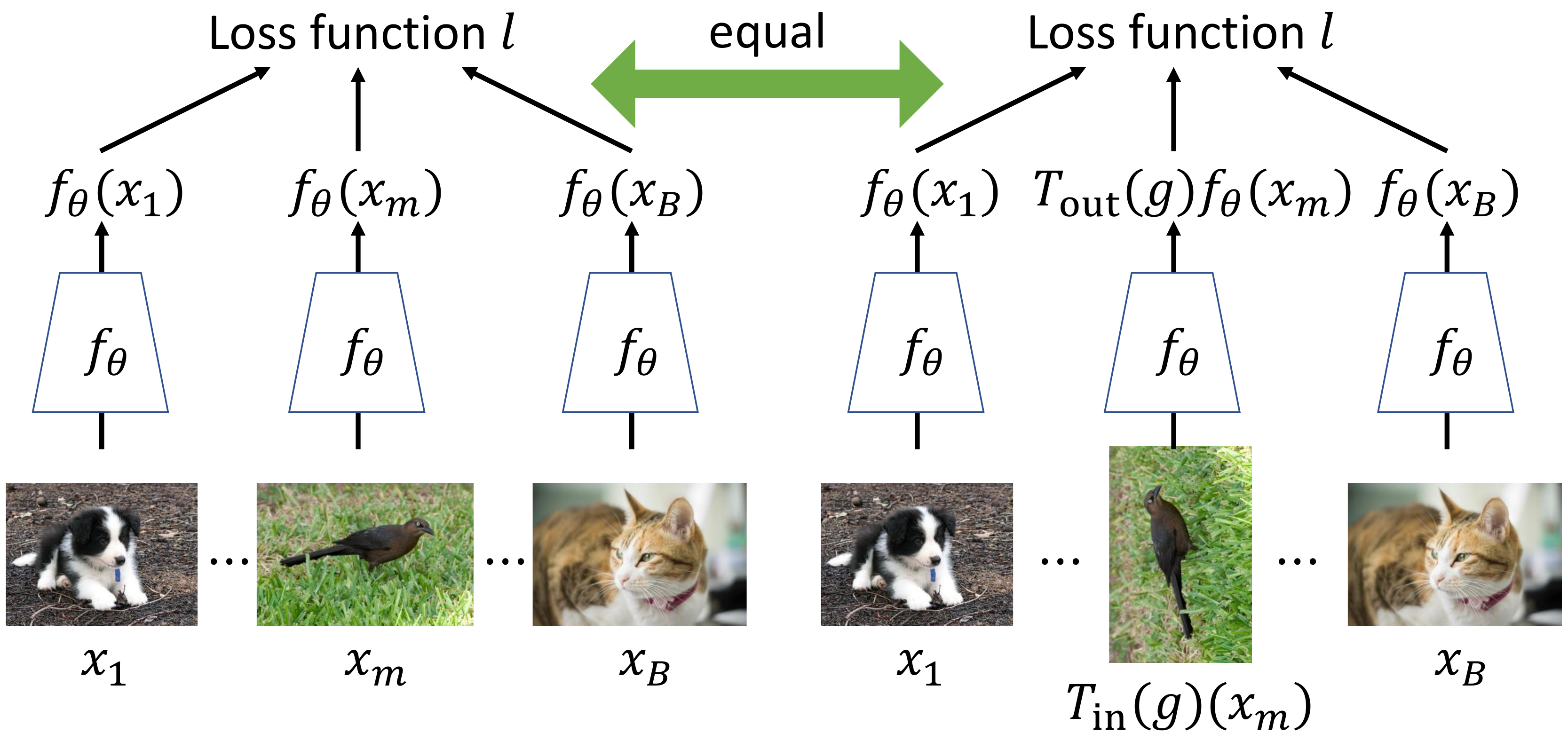}
  \caption{Illustrative image of invariant contrastive loss.}\label{fig:contrastive}
  \end{minipage}
\end{figure}

\subsubsection{Solving the jigsaw puzzle with $\pi/2$ rotation and image flipping}
Solving the jigsaw task \citep{noroozi2016unsupervised} is a task that extends context prediction.
We extract $3\times 3$ adjacent image subregions, apply the permutation to the subregions, and then input all permuted subregions to the classification model. The classifier is then trained to predict the permutation that was applied to the input subregions. When we consider all permutations, we need to learn a 9!-category classification problem, which is difficult to solve. Therefore, in general, we construct the subset whose permutations are distant from each other with respect to the Hamming distance and predict the permutation within the subset.
As shown in \figref{fig:jigsaw}, $\pi/2$ image rotation and image flipping act as the transformations on the permutation because these transformations themselves act as the permutation on the image subregions.
Therefore, by constructing the permutation subset such that it is closed under the effect of $\pi/2$ image rotation and image flipping, and aligning the permutation labels in the appropriate order, we can construct the label space that is equivariant under the transformations.
We denote the task using the equivariant permutation subset as an equivariant jigsaw.

\subsection{Invariant Contrastive Loss}
In the previous subsection, by assigning the equivariance to the label space as \equref{eq:pretext}, we can learn the model consistently even when we apply transformations to one of the input images.
We apply this idea to contrastive learning, where we use the relationship between the input images for training.
In contrastive learning, we first apply data augmentation on the input images to construct the input mini batch $\{x_b\}_{b=1}^B$, and then trains the model by minimizing $l(f_{\theta}(x_1),f_{\theta}(x_2),...,f_{\theta}(x_B))$, where $l$ is the predefined loss function.
When we apply a transformation to the input $x_m$ to use $T_{\mathrm{in}}(g)(x_m)$ as the input, using the equivariance of $f_{\theta}$, the loss function becomes $l(f_{\theta}(x_1),f_{\theta}(x_2),...,T_{\mathrm{out}}(g)f_{\theta}(x_m),...,f_{\theta}(x_B))$.
When the value of the loss function does not change under this transformation as
\begin{align}
& l(f_{\theta}(x_1),f_{\theta}(x_2),...,f_{\theta}(x_B)) \nonumber \\
  =& l(f_{\theta}(x_1),f_{\theta}(x_2),...,T_{\mathrm{out}}(g)f_{\theta}(x_m),...,f_{\theta}(x_B)),\label{eq:contrastive}
\end{align}
for any $m$, input $x_m$ and transformation $g$, the learned model that minimizes this loss is not affected by the transformations on the input.
We denote such losses that ignore the effect of the transformations on the output as the invariant contrastive loss.
The illustrative image of invariant contrastive loss is shown in \figref{fig:contrastive}.
Below, we apply the invariant contrastive loss to several tasks, including contrastive-based \citep{he2020momentum}, clustering-based \citep{caron2020unsupervised} and distillation-based \citep{chen2021exploring} tasks.

\subsubsection{Momentum Contrast}
Momentum Contrast \citep{he2020momentum} is the standard contrastive learning method. Momentum Contrast uses the learnable parameter $\theta$, momentum encoder parameter $\theta^k$, and memory bank $Q$ for training.
In each iteration, the model is trained as follows:
\begin{enumerate}
\item Sample minibatch $\{x_n\}_{n=1}^N$.
\item Apply data augmentation on the $\{x_n\}_{n=1}^N$ to obtain $\{\widehat{x^q_n}\}_{n=1}^N$ and $\{\widehat{x^k_n}\}_{n=1}^N$.
\item Apply $f_{\theta^k}$ on $\{\widehat{x^k_n}\}_{n=1}^N$ to obtain $\{k_n\}_{n=1}^N$.
\item Update $\theta$ with the loss
  \begin{equation}
    -\sum_{n=1}^N \frac{\exp\left(\frac{f_{\theta}(x^q_n)^t k_n}{\tau}\right)}{\exp\left(\frac{f_{\theta}(x^q_n)^t k_n}{\tau}\right) + \sum_{k\in Q} \exp\left(\frac{f_{\theta}(x^q_n)^t k}{\tau}\right) }. \label{eq:mocoloss}
  \end{equation}
\item Update $\theta^k$ using the moving average of $\theta$.
\item Update $Q$ with $\{k_n\}_{n=1}^N$ using first-in-first-out rule.
\end{enumerate}
We extend \equref{eq:mocoloss} to satisfy the condition of the invariant contrastive loss.
To this end, we modify the loss in order not to change even when we apply $T_{\mathrm{out}}(g)$ on $k$ and $f_{\theta}(x^q_n)$.
We focus on the inner product that appears in $\exp$ and replace it with
\begin{equation}
\frac{1}{|G|}\frac{1}{|G|}\sum_{g_1 \in G}\sum_{g_2 \in G}  (T_{\mathrm{out}}(g_1) f_{\theta}(x^q_n))^t  (T_{\mathrm{out}}(g_2) k).\label{eq:invinner}
\end{equation}
Because this value is the average of all pairs of transformations on the input features, the value of \equref{eq:invinner} does not change when we apply $T_{\mathrm{out}}(g)$ on $f_{\theta}(x^q_n)$ and $k$.
Therefore, we can construct the invariant contrastive loss using this inner product.
Further, using the bilinearity of the inner product, \equref{eq:invinner} can be calculated as the inner product between $\frac{1}{|G|} \sum_{g\in G} T_{\mathrm{out}}(g) f_{\theta}(x^q_n)$ and $\frac{1}{|G|} \sum_{g\in G} T_{\mathrm{out}}(g) k$.
Using this fact, the algorithm becomes invariant when we replace the feature extractor $f_{\theta}$ with its average $\frac{1}{|G|} \sum_{g\in G} T_{\mathrm{out}}(g) \circ f_{\theta}$, where $\circ$ denotes the composition of the function.
We call this method invariant Momentum Contrast.

\subsubsection{Swapping Assignments between Views}
We construct the invariant Momentum Contrast by replacing the inner product with the invariant inner product.
We can apply this procedure to the more complex contrastive learning method.
In this section, we apply our method to Swapping Assignments between Views (SwAV) \citep{caron2020unsupervised}.
SwAV trains the model with the loss that considers the consistency of the pseudo label assignments on the cluster centroids.
SwAV learns the feature extractor $f_{\theta}$ and the cluster centroids $C \in \mathbb{R}^{d\times c}$ where $d$ denotes the feature dimension and $c$ denotes the number of clusters as follows:
\begin{enumerate}
\item Sample minibatch $\{x_n\}_{n=1}^N$.
\item Apply data augmentation on $\{x_n\}_{n=1}^N$ to obtain $\{\widehat{x^q_n}\}_{n=1}^N$ and $\{\widehat{x^k_n}\}_{n=1}^N$.
\item Calculate the cluster assignment probability $q_n$ of $x_n$ using the entropy-regularized optimal transport between $\{f_{\theta}(\widehat{x^k_n})\}_{n=1}^N$ and $C$. We do not calculate the backpropagation through this part.
\item Calculate the predicted cluster assignment probability $p_n = \mathrm{softmax}(C^t f_{\theta}(\widehat{x^q_n}))$.
\item Update $\theta$ and $C$ with the cross-entropy loss $ - \sum_{n=1}^N q_n \log p_n$.
\end{enumerate}
The entropy-regularized optimal transport is solved by the iterative Sinkhorn-Knopp
algorithm \citep{cuturi2013sinkhorn}. This algorithm uses the cost function $C^t f_{\theta}(\widehat{x^k_n})$ and iteratively applies matrix-vector products to calculate the assignment.
Therefore, both $q_n$ and $p_n$ depend on the feature $f_{\theta}(x)$ as the form of $C^t f_{\theta}(x)$. We can construct the equivariant loss by replacing $C^t f_{\theta}(x)$ with the one that is invariant under the action of $T_{\mathrm{out}}(g)$.
Similar to invariant Momentum Contrast, we use $C^t \frac{1}{|G|} \sum_{g \in G} T_{\mathrm{out}}(g) f_{\theta}(x)$ as the invariant inner product. We denote this method as invariant SwAV.

\subsubsection{SimSiam}
SimSiam \citep{chen2021exploring} is a distillation-based method that trains the model to make the output features from the same input image close as follows:
\begin{enumerate}
\item Sample minibatch $\{x_n\}_{n=1}^N$.
\item Apply data augmentation on $\{x_n\}_{n=1}^N$ to obtain $\{\widehat{x^1_n}\}_{n=1}^N$ and $\{\widehat{x^2_n}\}_{n=1}^N$.
\item Apply $f_{\theta}$ on $\{\widehat{x^m_n}\}_{n=1}^N$ to obtain $\{z^m_n\}_{n=1}^N$ for $m=1,2$.
\item Apply predictor $h$ on $\{\widehat{z^m_n}\}_{n=1}^N$ to obtain $\{p^m_n\}_{n=1}^N$ for $m=1,2$.
\item Update $h$ and $f_{\theta}$ with the cosine similarity loss $ - \sum_{n=1}^N \frac{(z_n^1)^t p_n^2}{\|z_n^1\|\|p_n^2\|} + \frac{(z_n^2)^t p_n^1}{\|z_n^2\|\|p_n^1\|}$ while backpropagation through $z^m_n$ is ignored.
\end{enumerate}
In most cases, we can assume that $T_{\mathrm{out}}(g)$ preserves the L2 norm of the feature. For example, $T_{\mathrm{out}}(g)$ in G-CNNs act as the permutation between the feature elements, which can be represented as the orthogonal matrix. With this assumption, we simply need to replace the inner product $z_n^t p_n$ with the invariant one.
As in the case of Momentum Contrast, we conduct this by replacing $z^m_n$ with $\frac{1}{|G|} \sum_{g\in G} T_{\mathrm{out}}(g) z^m_n$ and $p^m_n$ with $\frac{1}{|G|} \sum_{g\in G} T_{\mathrm{out}}(g) p^m_n$ when we calculate the cosine similarity loss.
We call this method invariant SimSiam.

\begin{table*}[t]
  \caption{Accuracy (\%) on ImageNet with linear image classification setting.}
  \label{tab:imagenet}
  \centering
  \begin{tabular}{llll}
    \toprule
    Method & Baseline & Equivariant Model \& Loss (Ours) & Equivariant Model Only \\ \hline\hline
    Context prediction &32.7 & \textbf{35.1} &31.5\\
    Jigsaw &35.1 &\textbf{43.1} & 42.5 \\
    Momentum Contrast &63.8 &\textbf{65.7}&65.0 \\
    SwAV &71.4 &\textbf{71.6} & 68.2 \\
    SimSiam & 65.9 & \textbf{68.2} & 65.5 \\
    \bottomrule
  \end{tabular}
\end{table*}

\begin{table*}[t]
  \caption{Mean AP (\%) on VOC2007 with linear image classification setting.}
  \label{tab:voc2007}
  \centering
  \begin{tabular}{llll}
    \toprule
    Method & Baseline & Equivariant Model \& Loss (Ours) & Equivariant Model Only \\ \hline\hline
    Context prediction &51.7 &\textbf{53.6}  &49.1 \\
    Jigsaw &52.9 &56.7 &\textbf{57.8}  \\
    Momentum Contrast &80.7 &\textbf{81.1} &80.2 \\
    SwAV & 85.6 & \textbf{86.8}& 86.6 \\
    SimSiam & \textbf{81.7} & 81.1 & 81.5 \\
    \bottomrule
  \end{tabular}
\end{table*}

\begin{table*}[t]
  \caption{Accuracy (\%) on iNaturalist18 with linear image classification setting.}
  \label{tab:inaturalist18}
  \centering
  \begin{tabular}{llll}
    \toprule
    Method & Baseline & Equivariant Model \& Loss (Ours) & Equivariant Model Only \\ \hline\hline
    Context prediction &\textbf{8.56} & \textbf{8.56} &6.97 \\
    Jigsaw &8.72 &\textbf{13.8} &13.2  \\
    Momentum Contrast &33.4 &\textbf{33.8} &31.2 \\
    SwAV &\textbf{42.1} &35.8 &32.4  \\
    SimSiam & 32.6 & \textbf{33.7} & 27.8 \\
    \bottomrule
  \end{tabular}
\end{table*}

\begin{table*}[t]
  \caption{Accuracy and mean average precision (\%) with different transformation groups using Momentum Contrast.}
  \label{tab:ablation}
  \centering
  \begin{tabular}{l|llll}
    \toprule
    &Dataset & $\pi$ and Image Flipping & $\pi/2$ and Image Flipping & $\pi/4$ and Image Flipping \\ \hline\hline
    \multirow{3}{*}{Non Equivariant loss}&ImageNet & 64.6 &65.0 &63.8 \\
    &VOC2007 &79.8 & 80.2& 80.9 \\
    &iNaturalist & 32.4 & 31.2& 29.1 \\\hline
    \multirow{3}{*}{Equivariant loss (Ours)}&ImageNet &64.7 &65.7 &65.2 \\
    &VOC2007 &79.9 &81.1 & 80.4 \\
    &iNaturalist & 32.4& 33.8&30.9 \\
    \bottomrule
  \end{tabular}
\end{table*}

\begin{table*}[t]
  \caption{Accuracy and mean average precision (\%) with different transformation groups using SwAV.}
  \label{tab:ablationswav}
  \centering
  \begin{tabular}{l|llll}
    \toprule
    &Dataset & $\pi$ and Image Flipping & $\pi/2$ and Image Flipping & $\pi/4$ and Image Flipping \\ \hline\hline
    \multirow{3}{*}{Non Equivariant loss}&ImageNet & 68.5 &68.2 & 68.6 \\
    &VOC2007 &86.3 & 86.6& 86.9 \\
    &iNaturalist & 34.5 & 32.4& 32.6 \\\hline
    \multirow{3}{*}{Equivariant loss (Ours)}&ImageNet &71.5 &71.6 &70.9 \\
    &VOC2007 &86.3 &86.8 & 86.4 \\
    &iNaturalist & 38.5& 35.8&32.8 \\
    \bottomrule
  \end{tabular}
\end{table*}

\begin{table*}[t]
  \caption{Accuracy and mean average precision (\%) with different transformation groups using SimSiam.}
  \label{tab:ablationsimsiam}
  \centering
  \begin{tabular}{l|lll}
    \toprule
    &Dataset & $\pi$ and Image Flipping & $\pi/2$ and Image Flipping  \\ \hline\hline
    \multirow{3}{*}{Non Equivariant loss}&ImageNet & 65.0 &65.5  \\
    &VOC2007 &81.3 & 81.5 \\
    &iNaturalist & 28.9 &27.8 \\\hline
    \multirow{3}{*}{Equivariant loss (Ours)}&ImageNet &67.3 &68.2 \\
    &VOC2007 &81.3 &81.1  \\
    &iNaturalist & 33.0& 33.7 \\
    \bottomrule
  \end{tabular}
\end{table*}

\section{Experiment}
In this section, we evaluate the proposed self-supervised loss on standard image recognition benchmarks.
\subsection{Experimental Setting}
\subsubsection{Dataset for pretraining}
We use the standard large-scale image recognition benchmark ImageNet \citep{deng2009imagenet} as the resource for self-supervised pretraining.
ImageNet is a general image recognition dataset that consists of approximately 1,300,000 images with 1,000 categories.
In the pretraining phase, we do not use the label information, and train the model with the self-supervised loss.
\subsubsection{Model architecture}\label{sec:arch}
We use ResNet50 \citep{he2016deep} as the baseline model.
We apply non-equivariant baseline self-supervised methods on this model.
For the group equivariant neural networks to learn with the proposed loss, we use the model that replaces each convolutional layer of ResNet50 with the corresponding equivariant layers (Eq. (\ref{eq:equivariantconv0}, \ref{eq:equivariantconv})).
To match the number of learnable parameters to the original ResNet50, we divide the number of filters by $\sqrt{|G|}$ from the original model.
The equivariant ResNet50 we use has almost the same number of learnable parameters as the original ResNet50 and the dimension of the features before the last fully-connected layer becomes $\sqrt{|G|}$ times the original feature dimension.
We also replace the projection head with the corresponding equivariant layers with the reduced number of feature dimensions except for the output feature.
We match the output feature dimension for the contrastive loss.
For the transformations $G$, we use a group that consists of $\pi/2$ rotation $(|G|=4)$ for the context prediction task, and a group that consists of $\pi/2$ rotation and image flipping $(|G|=8)$ for the other tasks.
\subsubsection{Comparison method}
As the baseline method, we apply context prediction \citep{doersch2015unsupervised}, jigsaw \citep{noroozi2016unsupervised}, Momentum Contrast \citep{he2020momentum}, SwAV \citep{caron2020unsupervised} and SimSiam \citep{chen2021exploring} on ResNet50.
As in the proposed method, we apply the proposed equivariant variants of these methods to the group equivariant ResNet50.

As an ablation study to evaluate the effect of the equivariant model architecture and the effect of the proposed equivariant loss separately, we also evaluate the performance of the model that we train the group equivariant ResNet50 with the existing non-equivariant self-supervised loss as follows.
We use a different order of the relative position labels from the order proposed in the previous section for context prediction, the original permutation subset for the jigsaw task and we use the original inner product to calculate the loss function for the Momentum Contrast, SwAV and SimSiam.

\subsubsection{Training setting}
We refer to \cite{chen2021exploring} for training SimSiam.
We refer to \cite{goyal2019scaling} and its extension \cite{goyal2021vissl} for the other methods.

For context prediction, we use a two-layer mlp consisting of a layer that reduces the feature dimension from 2,048 to 1,000, and a layer that maps the 2,000-dimensional feature consisting of two 1,000-dimensional subregion features to eight-dimensional category space.
We use random grayscaling as the data augmentation and train the model for 105 epochs with a minibatch size of 256.
We use momentum SGD for training. We use linear warm up to increase the learning rate from 0.025 to 0.1 and then multiply by 0.1 on 30, 60, 90, and 100 epochs.
We set the momentum as 0.9 and the weight decay rate as 0.0001.

For the jigsaw task, we use the subset consisting of 2,000 permutations and train the model as a 2,000-way classification problem.
Similar to the case of context prediction, we use a two-layer mlp. The first layer reduces the dimension from 2,048 to 1,000, and the succeeding layer reduces the 9,000-dimensional feature consisting of nine 1,000-dimensional subregion features to a 2,000-dimensional category space. We use the same learning parameter as that for the context prediction task.

For the Momentum Contrast, we use the module introduced in Momentum Contrast v2 \citep{chen2020improved} and use the two-layer mlp with 2,048 mid-feature dimension and 128 output feature dimension as the projection head.
We use random cropping, random color distortion, random Gaussian blur and random image flipping as the data augmentation.
We set the size of the memory bank as 65,536, the temperature parameter as 0.2, and the coefficient of moving average for the momentum encoder as 0.999.
We train the model for 200 epochs with batch size 256. We set the momentum as 0.9 and the weight decay rate as 0.001. We initialize the learning rate with 0.03 and multiply by 0.1 at 120 and 160 epochs.

For SwAV, following the original paper, we use $2\times 224 + 6\times 96$ multi-crop as the input.
We use random cropping, random color distortion, random Gaussian blur, and random image flipping as the data augmentation.
We use the same architecture as Momentum Contrast for the projection head and set the number of clusters as 3,000.
We set the batch size as 256 and train the model for 200 epochs. We set the initial learning rate as 0.6 and use LARC to modify the learning rate.
We set the momentum as 0.9 and the weight decay rate as 0.000001.
We use the temperature parameter as 0.1 and the number of iterations as 3 for the iterative Sinkhorn-Knopp algorithm.
Further, following the original paper, we use a queue with a length of 3,840 after 15 epochs and also use samples from a queue when we calculate the assignment.

For SimSiam, we use a two-layer mlp with 2,048 mid-feature dimension and 512 output feature dimension as the projection head.
We use random cropping, random color distortion, random gray scale, random Gaussian blur, tand random image flipping as the data augmentation.
We train the model for 100 epochs with the batch size 512. We set the momentum as 0.9 and the weight decay rate as 0.0001. We initialize the learning rate with 0.05 and adjust the learning rate with a cosine decay schedule.

As described in \secref{sec:arch}, we replace the projection head with the equivariant layers with the same output feature dimension for the proposed loss.
We use the same learning parameter for the equivariant and non-equivariant methods.

\subsubsection{Evaluation Setting}
We evaluate the pretrained model using the standard image recognition benchmarks.
We evaluate the model using a linear image classification setting in which we fix the pretrained model and only learn the last fully-connection layer on top of the pretrained network.
In addition to the ImageNet that is also used for the pretraining, we use the PASCAL VOC 2007 (VOC2007) \citep{Everingham15} and iNaturalist18 \citep{van2018inaturalist} datasets.
VOC2007 is a traditional image recognition benchmark that consists of approximately 2,500 images for training and evaluation with 20 categories.
iNaturalist18 dataset is the fine-grained image recognition dataset that consists of approximately 450,000 images with approximately 8,000 categories of animal species.

For ImageNet, following \cite{goyal2019scaling} context prediction and jigsaw tasks, we apply the average pooling with kernel size $6\times 6$ at the four corners of the output of the last convolutional layer with size $7\times 7\times 2,048 \sqrt{|G|}$ and align to obtain $8,192 \sqrt{|G|}$-dimensional features. We then apply a linear classification layer to predict the category.
For Momentum Contrast and SwAV, following the original paper, we apply global average pooling and apply linear classification on the $2,048 \sqrt{|G|}$-dimensional features.
We train the model for 28 epochs with a batch size of 256 for the context prediction and jigsaw task. We set the weight decay rate as 0.0005 and momentum as 0.9.
We set the base learning rate as 0.01 and multiply by 0.1 on the 8, 16, and 24 epochs.
For Momentum Contrast, we set the batch size as 256, weight decay rate as 0, momentum as 0.9, and base learning rate as 30.0. We multiply the learning rate by 0.1 on 60 and 80 epochs and train the model for 100 epoch.
For SwAV, we set the batch size as 256, weight decay rate as 0.000001, momentum as 0.9, and base learning rate as 0.3. We modify the learning rate with cosine learning rate decay and train the model for 100 epochs.
For SimSiam, we set the batch size as 4,096, weight decay rate as 0, momentum as 0.9, and base learning rate as 0.1. We apply LARC for the learning rate modification. We train the model for 90 epochs.

For VOC2007, we apply the linear SVM on the output of global average pooling.

For iNaturalist18, we set the batch size as 256, weight decay rate as 0.0005, and base learning rate as 0.001. We train the model for 84 epochs with the learning rate multiplied by 0.1 on 24, 48, and 72 epochs.

\subsubsection{Implementation}
We implemented the self-supervised methods using \cite{goyal2021vissl} and implemented the G-CNNs using \cite{e2cnn}.
We used eight V100 or A100 GPUs for pretraining and one for fine-tuning the models.
We conducted the experiment for both the existing methods and the proposed method and report the score in our setting.
Therefore, there exist cases for which the scores of the existing methods are different from those reported in previous studies.

\subsection{Results}
Tables \ref{tab:imagenet}, \ref{tab:voc2007} and \ref{tab:inaturalist18} show the results.
In many settings, the proposed method demonstrates better accuracy than the existing non-equivariant baselines.
This indicates that group equivariant neural networks have higher representation ability than the standard non-equivariant models even in the self-supervised setting.
When we compare the accuracy using the equivariant CNNs with the proposed and existing losses, there exist cases for which the equivariant model with non-equivariant loss demonstrate lower accuracy than the original non-equivariant model.
This implies the effectiveness of combining the proposed equivariant loss with the equivariant model.
For the jigsaw case, the performance drop when using non-equivariant loss is relatively small compared to context prediction and in some cases non-equivariant loss demonstrate better performance.
Because jigsaw solves 2,000-way classification compared to eight-category classification for context prediction, we expect that the effect of equivariance on the label space becomes relatively small.
Moreover, when we use contrastive loss, the model that combines G-CNNs with non-equivariant loss often demonstrates lower accuracy than the non-equivariant model.
This implies that applying the contrastive loss directly to the equivariant model may diminish the performance.
These results show the effectiveness of the proposed equivariant loss when we train the equivariant networks.

\subsection{Ablation Study on the Transformation Group}
The previous experiment mainly handles the equivariance for the group that consists of $\pi/2$ image rotation and image flipping.
For the case of invariant contrastive loss, we can use any group whenever we can calculate the average feature.
For the ablation study, we evaluated the model that applies the different transformation groups to the invariant Momentum Contrast, invariant SwAV and invariant SimSiam.
For the transformation group, in addition to the group consisting of $\pi/2$ rotation and image flipping, we used the group consisting of $\pi$ rotation and image flipping $(|G|=4)$ and the group consisting of $\pi/4$ rotation and image flipping $(|G|=16)$. We omitted the SimSiam with the group consisting of $\pi/4$ rotation and image flipping $(|G|=16)$ owing to the memory limitation.
We used the same dataset and training setting and evaluated the accuracy and mean average precision.

Tables \ref{tab:ablation}, \ref{tab:ablationswav} and \ref{tab:ablationsimsiam} show the results.
There is a tendency that the group that considers $\pi/2$ rotations demonstrates the best performance among the transformation groups, which implies that considering appropriate transformations contributes to the performance.
Regarding the comparison between the proposed and non-equivariant losses, the proposed method demonstrates better accuracy in most settings.
This indicates that the equivariant loss is more effective even when we change the group size.

\section{Conclusion}
In this study, we proposed a method to construct the loss for training group equivariant neural networks in an unsupervised manner.
To train the model that is robust under the transformation of the input data, we aim for a loss function that is invariant under the transformation.
To construct such an invariant loss, we propose the concept of equivariant pretext labels and invariant contrastive loss.
We then proposed the equivariant versions for several existing self-supervised methods.
Experiments on standard image recognition benchmarks demonstrate that we can obtain good pretrained models by combining the proposed loss with the equivariant neural networks.
We expect that we can apply the method of equivariant pretext labels and invariant contrastive loss to wider self-supervised tasks, which will be pursued in future work.

\section*{Acknowledgements}
This work was partially supported by JSPS KAKENHI Grant Number JP19K20290, JST Moonshot R\&D Grant Number JPMJPS2011, CREST Grant Number JPMJCR2015 and Basic Research Grant (Super AI) of Institute for AI and Beyond of the University of Tokyo.

\bibliographystyle{unsrtnat}
\bibliography{aaai23}

\end{document}